\pdfoutput=1
\documentclass{article}

 \usepackage[nonatbib,final]{neurips_2022_ml4ps}

\usepackage[utf8]{inputenc} % allow utf-8 input
\usepackage[T1]{fontenc}    % use 8-bit T1 fonts
\usepackage{url}            % simple URL typesetting
\usepackage{booktabs}       % professional-quality tables
\usepackage{amsfonts}       % blackboard math symbols
\usepackage{nicefrac}       % compact symbols for 1/2, etc.
\usepackage{microtype}      % microtypography
\usepackage{subfiles}
\usepackage{svg}
\usepackage{float}
\usepackage{adjustbox}
\usepackage{amsmath}

\title{Continual learning autoencoder training for a particle-in-cell simulation via streaming}

\author{%
  Patrick Stiller \\
  Institute for Radiation \\
  Helmholtz-Zentrum Dresden-Rossendorf\\
  \texttt{p.stiller@hzdr.de} \\
   \And
    Varun Makdani\\
  Institute for Radiation \\
  Helmholtz-Zentrum Dresden-Rossendorf\\
  \texttt{v.makdani@hzdr.de} \\
     \And
  Franz Pöschel \\
  Center for Advanced Systems Understanding Görlitz\\
    Helmholtz-Zentrum Dresden-Rossendorf\\
  \texttt{f.poeschel@hzdr.de} \\
     \And
   Richard Pausch \\
  Institute for Radiation \\
  Helmholtz-Zentrum Dresden-Rossendorf\\
  \texttt{r.pausch@hzdr.de} \\
  \And
    Alexander Debus \\
  Institute for Radiation \\
  Helmholtz-Zenrum Dresden-Rossendorf\\
  \texttt{a.debus@hzdr.de} \\
    \And
  Michael Bussmann\\
  Center for Advanced Systems Understanding Görlitz\\
  Helmholtz-Zenrum Dresden-Rossendorf\\
  \texttt{m.bussmann@hzdr.de} \\
    \And
   Nico Hoffmann \\
  Institute for Radiation \\
  Helmholtz-Zentrum Dresden-Rossendorf\\
  \texttt{n.hoffmann@hzdr.de} \\
}

\begin{document}

\maketitle

\begin{abstract}
The upcoming exascale era will provide a new generation of physics simulations. These simulations will have a high spatiotemporal resolution, which will impact the training of machine learning models since storing a high amount of simulation data on disk is nearly impossible. Therefore, we need to rethink the training of machine learning models for simulations for the upcoming exascale era. This work presents an approach that trains a neural network concurrently to a running simulation without storing data on a disk. The training pipeline accesses the training data by in-memory streaming. Furthermore, we apply methods from the domain of continual learning to enhance the generalization of the model. 
We tested our pipeline on the training of a 3d autoencoder trained concurrently to laser wakefield acceleration particle-in-cell simulation. Furthermore, we experimented with various continual learning methods and their effect on the generalization. 
\end{abstract}

\section{Introduction}
Numerical simulations of complex systems such as Laser-Plasma acceleration are computationally costly and run on the largest HPC systems in the world. Neural Network based surrogate models of these simulations drastically speeds up the analysis due to fast inference times promising in situ analysis of experimental data.  Such surrogate models can be applied for forward and inverse problems various fields of science, such as quantum physics \cite{neuralsolvers}, medical science \cite{thermalprocess}, and plasma physics \cite{plasmasurrogate}. High-fidelity simulations in the upcoming exascale era can provide high-resolution training data for surrogate models. However, providing a training dataset stored on a disk is nearly impossible for these simulations. Therefore, we need to rethink the training of surrogate models to tackle the memory- and space limitations of current HPC systems. We conducted in-memory coupling to access the training data sequentially without storing them on disk. In-memory coupling allows streaming simulation data from multiple compute nodes at network speeds. However, in such a streaming setup, data, once consumed, cannot be regained. In such a learning environment, avoiding catastrophic forgetting is a big challenge.
In this work, we showcase the prototype of this novel training pipeline. We applied techniques from continual learning into our streaming training pipeline to foster the generalization of the trained model and avoid catastrophic forgetting. We have experimented with various regularization and replay-based methods to avoid forgetting in a continuous learning regime. Such as A-GEM \cite{a-gem} and Online EWC \cite{ewc}. We tested our pipeline on time-dependent 3D-electric field data generated from an LWFA (laser wakefield acceleration) simulation on PIConGPU \cite{picONgpu}. Additionally, we extended the A-GEM algorithm for autoencoder training to reduce the memory consumption of this promising approach. This framework for continual learning is developed in an extensible manner, enabling its ability to be applied to model other simulation data. 
The paper is structured as follows. In section 2, we describe our conducted method. We discuss the results, the effect of different continual learning methods in our pipeline, and their memory consumption in section 3. Section 4 contains the concluding remarks.
\begin{figure}[H]
    \centering
    \includegraphics[width=\linewidth]{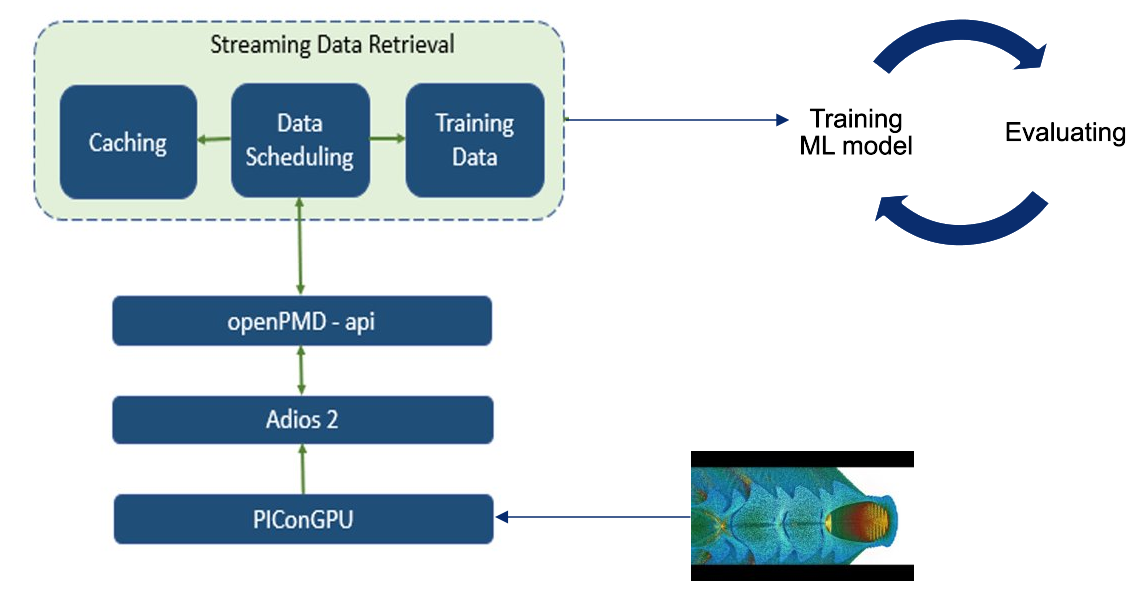}
    \caption{Illustration of the training pipeline. We access our electrical field data from a PIConGPU simulation via the Adios2 Backend of the openPMD-api. Additionally, we added a data scheduling unit to prevent the simulation from stopping }
    \label{fig:pipeline}
\end{figure}
\section{Method}
We developed a training pipeline that allows in-memory access via streaming to the simulation data while the simulation is running (Figure \ref{fig:pipeline}). The Adios2 library allows access to the simulation data via in-memory streaming. Our OpenPMD unit converts the streamed Adios2 data into an OpenPMD data frame which allows our training engine to access the training data. The training engine consists of a data scheduler, which stops and starts the simulation and stores intermediate results in a cache. Caching the data reduces the number of stops in the simulation. The neural network uses the full cache for the next training iteration. 

The disadvantage of this pipeline is that the trained model cannot revisit simulation data once consumed. Therefore, we applied continual learning techniques to improve the generalization and robustness of the trained model. We performed a 3d autoencoder training to learn a low-dimensional representation of the underlying simulation data and apply it for subsequent simulations. 
We have proposed a method, inspired by the replay-based method A-GEM\cite{a-gem}, that significantly reduces the memory requirement for autoencoder training. The basic idea of the A-GEM algorithm is a projection of the current gradient $g$ to a reference gradient $g_\mathrm{ref}$ to avoid catastrophic forgetting. The reference gradient is computed with respect to input data $m$, which is sampled from episodic memory $\mathcal{M}$ (see equation \ref{eq:gref}). The projection only will be applied if $\tilde{g}^{\top} g_{r e f} \geq 0$
is fulfilled.
\begin{equation} \label{eq:gref}
    \tilde{g} = g -\frac{g^\mathsf{T}g_{\mathrm{ref}}}{g_{\mathrm{ref}}^\mathsf{T}g_{\mathrm{ref}}}g_{\mathrm{ref}}
\end{equation}

However, applying an episodic memory will be too expensive for high-fidelity simulation data. Therefore, we conducted a method that exploits the low dimensionality of the latent space of the autoencoder and stores them instead in the episodic memory. Hence it allows using A-GEM for such massive data as experienced in the use-case of LWFA simulation, especially for the upcoming exascale era. The loss for reference-gradient ($g_{\mathrm{ref}}$) can then be calculated based on the change in the reconstruction of latent space by randomly sampling encoded reference memory from memory replay and then passing it through the decoder and then again with the encoder. 

\begin{equation}
    g_{ref} = \nabla_\theta \|f_E(f_D(e))- e\|^2_2  \quad  \textrm{ where }  e \subset \cup_{k<t} \mathcal{E}_{k}
\end{equation}

here $f_E(f_D(e))$ is the encoder output on input data $e$ sampled from replay memory ($\mathcal{E}$) containing samples of learned encoded latent space representations of previously learned tasks. 
Additionally, we adapted the method for convolutional autoencoders. Every convolutional layer extracts different features using different filters and on various output channels. In the case of A-GEM, layers with a large magnitude of gradients influence the dot product for the condition of parameter constraint ($\tilde{g}^{\top} g_{r e f} \geq 0$), even the layers which do not require change are affected by other layers, which intuitively may increase loss on the objective. This is the basis of the second modification, where the condition is checked layer-wise, and the gradients are projected layer-wise as well.
\section{Results}
 We trained a 3d convolutional autoencoder on the electrical field which is streamed sequentially from a PIConGPU \cite{picONgpu} LWFA simulation via the Adios2-backend [\cite{adios2}] of OpenPMD \cite{openPMD}.
Each simulation step denotes a training task for the autoencoder. The proposed neural network architecture is inspired by \cite{deepfluid} and \cite{shadowgraphy} reduced order models for fluid simulations and shadowgraphy. We integrated the Online-EWC \cite{ewc} and the A-GEM algorithm \cite{a-gem} into our training pipeline to compare their effect on the generalization. We used an episodic memory size of 10 for A-GEM and our proposed method. The training and the simulations run on a Tesla V100. We trained 40 epochs per streamed electrical field and used the Adam optimizer \cite{adam}.
 
\begin{figure}[H]
        \centering
        \includegraphics[width=\textwidth]{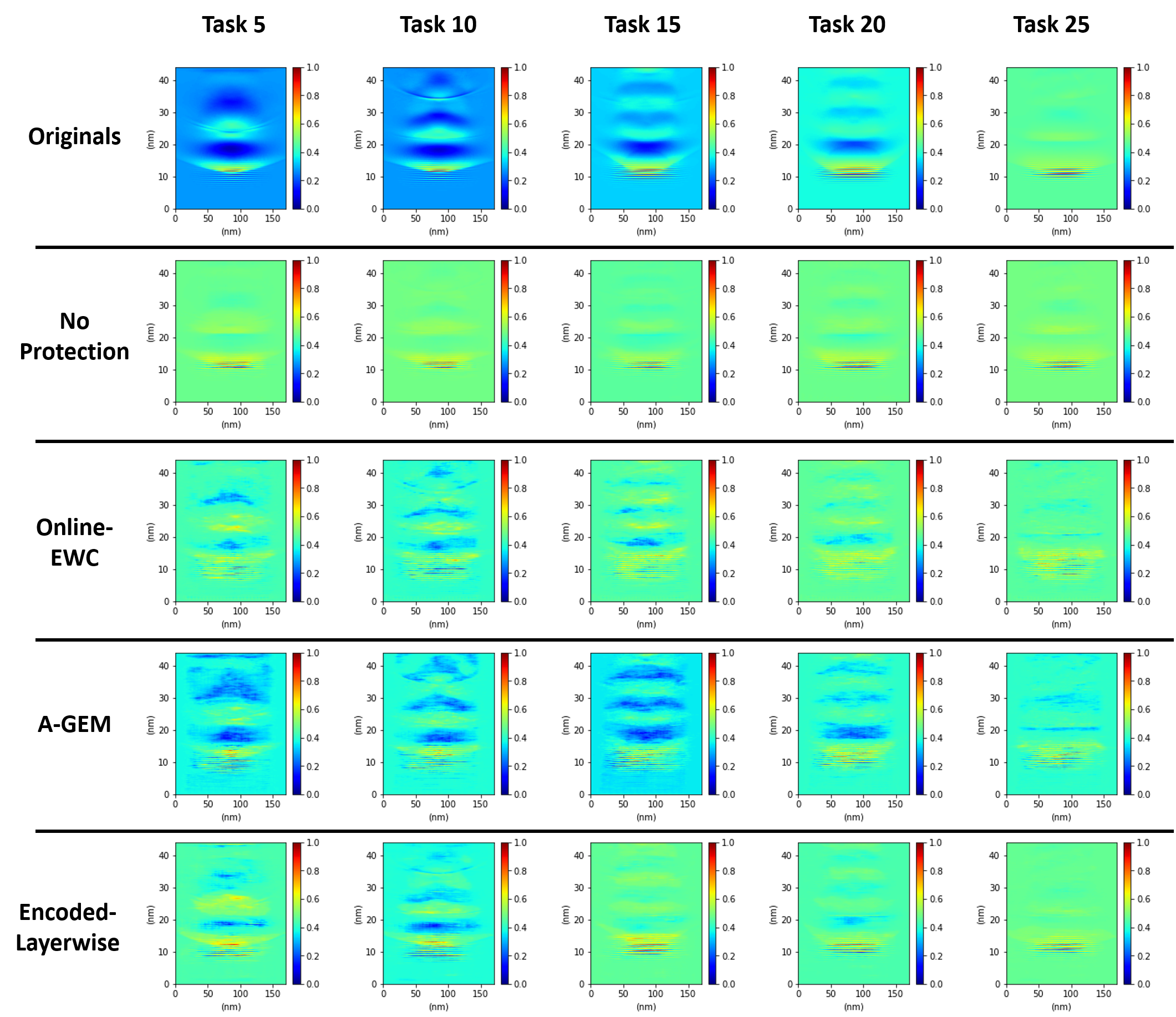}
        \caption{Post-training reconstructions of electrical fields by autoencoders trained with different continual learning techniques.The first row visualizes the ground truth electrical field. Each task corresponds to a simulation step}
        \label{fig:online-recon-image-grid}
    \end{figure}

 Table \ref{table:all-tasks-metrics} compares the post-training performance of Online-EWC, A-GEM, and our proposed method. We used the average L1-Norm to measure the pixel-wise similarity. Furthermore, we used the SSIM norm \cite{ssim} to measure structural similarity. Additionally, we performed basic training to quantify the effect of the continual learning techniques. Figure \ref{fig:online-recon-image-grid} visualizes the electrical fields after a full training beside a simulation. We observed that the autoencoder memorizes, as expected, only the most recent electrical field if he got trained without catastrophic forgetting protection. The Online-EWC algorithm learns a mixture of seen states which has no improvement in reconstruction quality. The A-GEM algorithm performs the best in terms of reconstruction quality (pixel-wise and structural similarity), followed by our proposed method. However, all methods cannot reconstruct the high-oscillating patterns in the electrical field data. We need more research to make the autoencoder robust for that pattern.
 
 Table \ref{table:all-tasks-metrics} additionally compares the memory overhead for the proposed methods. Since A-GEM stores ten electric fields simultaneously, we measured the highest additional memory consumption. The A-GEM method would not be feasible for use cases with higher resolution of the electric fields. The additional memory usage of the online EWC highly depends on the model size since it stores the fisher information of the model of the previous simulation step. Our proposed method provides the lowest additional memory usage because it only stores latent representations of the previous time steps.

\begin{table}[H]
\centering
\begin{tabular}{lccc}
\hline
Method              & L1 - Loss & SSIM  & Memory Overhead\\ \hline
No Protection                           & 0.16      
& 0.54    & 0 MB  \\ 
Online - EWC                            & 0.19               & 0.18   &  94 MB     \\ 
A-GEM                                   & \textbf{0.11}      & \textbf{0.59} & 2400 MB \\ 
Encode-Layerwise (Proposed)      & 0.13               & 0.46        &  \textbf{0.03 MB} \\ \hline
\end{tabular}
\caption{Comparison of the autoencoder performance trained with different continual learning techniques. All values are the average values through all simulation time-steps. Additionally, we considered the memory overhead of each method}
\label{table:all-tasks-metrics}
\end{table}

\section{Conclusion}
In this work, we presented a novel training pipeline for training an autoencoder concurrently to a running simulation while accessing the training data via streaming. We applied various continual learning techniques to improve the autoencoder's robustness and generalization and reduce the effect of catastrophic forgetting. We compared the applied methods with respect to reconstruction error, recovering structures, and memory consumption. Additionally, we extended the A-GEM algorithm for autoencoder training which heavily decreased the memory consumption of A-GEM. We conclude that replay-based methods, such as A-GEM and our proposed method, provide more accurate reconstruction since they access older simulation steps. Unfortunately, the trained autoencoder is only able to recover some high-oscillating patterns when it is trained in an online manner. We need to do more research to improve the trained autoencoder's reconstruction quality. In the future, we want to apply our approach on a larger scale and more sophisticated simulations.
\bibliographystyle{unsrt}
\bibliography{main}
\end{document}